\documentclass{article}

\usepackage{arxiv}
\usepackage{float}
\usepackage[utf8]{inputenc} 
\usepackage[T1]{fontenc}    
\usepackage{hyperref}       
\usepackage{url}            
\usepackage{graphicx}
\usepackage{multirow}
\usepackage{booktabs}       
\usepackage{flafter}		
\usepackage{longtable}
\usepackage{plantuml}
\usepackage{acronym}
\usepackage[acronym]{glossaries}

\newacronym{LLMs}{LLMs}{Large Language Models}
\newacronym{PII}{PII}{Personal Identifiable Information}
\newacronym{UPT}{UPT}{User Provided Tokens}
\newacronym{NER}{NER}{Named Entity Recognition}
\newacronym{PoS}{PoS}{Part of Speech}
\newacronym{IL}{IL}{Information Loss}
\newacronym{ILM}{ILM}{Manual information loss}
\newacronym{ILS}{ILS}{Similarity based Information loss}
\newacronym{TT}{TT}{Transformation Technique}
\newacronym{STT}{STT}{Sensitivity to Transformation Technique}

\graphicspath{ {./images/} }
\title{ Life of PII – A PII Obfuscation Transformer }

\date{} 					

\author{ 
    \href{https://orcid.org/0009-0004-1722-0306}
    {\includegraphics[scale=0.06]{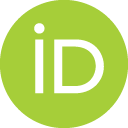}\hspace{1mm}Ajinkya Deshmukh}\\
	Synechron Data Practice (Pune, India)\\
    \texttt{Ajinkya.Deshmukh@synechron.com}\\
    \And
    \href{https://orcid.org/0000-0003-1002-1828}
    {\includegraphics[scale=0.06]{orcid}\hspace{1mm}Saumya Banthia}\\
	Synechron Data Practice (Charlotte, US)\\
    \texttt{Saumya.Banthia@synechron.com}\\
    \And
    \href{https://orcid.org/0000-0002-9064-3362}
    {\includegraphics[scale=0.06]{orcid}\hspace{1mm}Anantha Sharma}\\
	Synechron Data Practice (Charlotte, US)\\
    \texttt{Anantha.Sharma@synechron.com}
}

     \renewcommand{\headeright}{Life of PII}
 \renewcommand{\undertitle}{Synechron Data Practice}

\begin{document}
\maketitle


\begin{abstract}
Protecting sensitive information is crucial in today's world of \gls{LLMs} and data-driven services. One common method used to preserve privacy is by using data perturbation techniques to reduce overreaching utility of (sensitive) \gls{PII} data while maintaining its statistical and semantic properties. Data perturbation methods often result in significant information loss, making them impractical for use. In this paper, we propose 'Life of \gls{PII}', a novel Obfuscation Transformer framework for transforming \gls{PII} into faux-\gls{PII} while preserving the original information, intent, and context as much as possible. Our approach includes an API to interface with the given document, a configuration-based obfuscator, and a model based on the Transformer architecture, which has shown high context preservation and performance in natural language processing tasks and \gls{LLMs}. 

Our Transformer-based approach learns mapping between the original \gls{PII} and its transformed faux-\gls{PII} representation, which we call "obfuscated" data. Our experiments demonstrate that our method, called Life of \gls{PII}, outperforms traditional data perturbation techniques in terms of both utility preservation and privacy protection. We show that our approach can effectively reduce utility loss while preserving the original information, offering greater flexibility in the trade-off between privacy protection and data utility. Our work provides a solution for protecting \gls{PII} in various real-world applications. 
\end{abstract}

\section{Introduction}

The Use of \gls{LLMs} like Chat-GPT is increasing in the world, people from all backgrounds have started to use these tools to keep at pace with the world and to make their lives easier. Many financial, insurance companies and investment banks want to use these \gls{LLMs} to satisfy their requirements. One of the main challenges which they are facing is maintaining their data privacy because these \gls{LLMs} are not yet hostable on their private servers. Most financial institution's data is rich in \gls{PII} and therefore carries with it an extra risk, sending it out. One solution to this problem is to transform and obfuscate this data prior to sending it to the \gls{LLMs} like Chat-GPT, get the response from the \gls{LLMs} and then again re-transform this obfuscated response to get the final answer. This is what our approach discusses as part of this paper.

In a previous work, we had explored transformer-based models for use in Question Answering \cite{qnaPaper}. This gives us the opportunity and scope to provide a solution to this problem. Our approach provides transforming and re-transforming of data within the organization maintaining data privacy. It also ensures that transformed and obfuscated data retains its semantic meaning. This paper focuses on the use of python dependency modules and Natural Language Processing techniques to provide transforming and re-transforming facilities.

\section{Methodology}
To transform the data, three transformation techniques were implemented which are \gls{UPT} Transformation, \gls{NER} Transformation and \gls{PoS} Transformation and checked response from \gls{LLMs} for different combinations of these transformation techniques. Flow for text transformation and LLM question and answering system is as below: 

\begin{figure}[ht]
    \centering
    \includegraphics[width=0.75\textwidth]{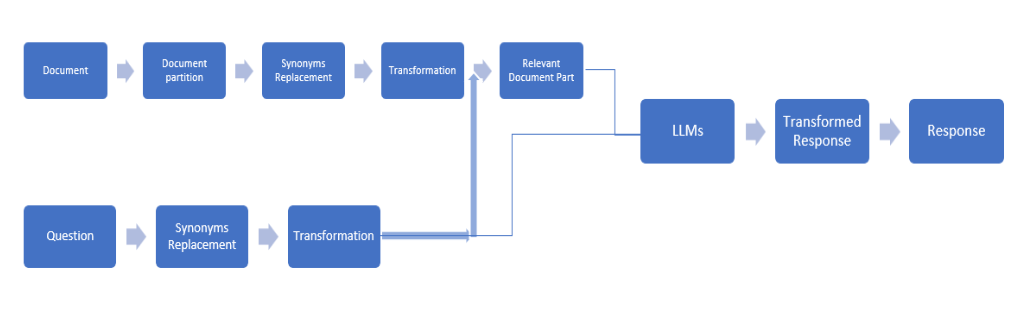}
    \caption{Proposed Text Transformation Flow for \gls{LLMs} Question and Answering}
    \label{fig:mesh1}
\end{figure}

The step for synonyms replacement is carried out to ensure transformation of that word and also to avoid different transformation for the same word. Transformation techniques used are as below: 

\subsection{{UPT} Transformation}
\gls{UPT} Transformation consists of providing words (which user wants to hide) inside configuration and providing tokens for those words. Then after applying \gls{UPT} Transformation, these words will be replaced with the tokens user has provided. For example, when user provides token ‘D202’ for the word ‘Krypton’ and token ‘Meridian’ for word ‘Eastern Richard’. Result of this transformation will be: 

\begin{table}[h]
\centering
\begin{tabular}{|p{3in}|p{3in}|}
\hline
Original Text  & The \textbf{Eastern Richard} Company Monthly Status Report states that it is performing good, but Project \textbf{Krypton} has a red status.  \\ \hline
\gls{UPT} Transformed Text  & The \textbf{Meridian} Company Monthly Status Report states that it is performing good, but Project \textbf{D202} has a red status.  \\ \hline
\end{tabular}
\end{table}

As this transformation hid information user wanted to hide and at the same time ensured semantic meaning.  Word ‘Eastern Richard’ is hiding with word ‘Meridian’ and ‘Meridian’ is also keeping the semantic meaning as it sounds like one company.

\subsection{{NER} Transformation}
\gls{NER} Transformation consists of identifying if there are any named entities in text \cite{stanzapaper} and replacing these named entities with some tokens. Considering same example as above, result of \gls{NER} transformation will be: 

\begin{table}[h]
\centering
\begin{tabular}{|p{3in}|p{3in}|}
\hline
Original Text    & \textbf{The Eastern Richard Company Monthly Status Report} states that it is performing good, but \textbf{Project Krypton} has a red status.   \\ \hline
\gls{NER} Transformed Text   & \textbf{N0} states that it is performing good, but \textbf{N1} has a red status. .  \\ \hline
\end{tabular}
\end{table}

It can be observed that named entities found in the original text are ‘The Eastern Richard Company Monthly Status Report’ and ‘Project Krypton’ which were replaced with ‘N0’ and ‘N1’. 
\pagebreak

\subsection{{PoS} Transformation}
\gls{PoS} Transformation consists of identifying if there are parts of speech like nouns in the text \cite{stanzapaper} and replacing these nouns with some tokens. Considering same example as above, result of \gls{PoS} transformation will be: 

\begin{table}[h]
\centering
\begin{tabular}{|p{3in}|p{3in}|}
\hline
Original Text    & The \textbf{Eastern Richard Company Monthly Status} Report states that it is performing good, but \textbf{Project Krypton} has a red status.   \\ \hline
\gls{PoS} Transformed Text  & The \textbf{P0} Report states that it is performing good, but \textbf{P1} has a red status.   \\ \hline
\end{tabular}
\end{table}

It can be observed that \gls{PoS} I.e., nouns found in the original text are ‘Eastern Richard Company Monthly Status’ and ‘Project Krypton’ which were replaced with ‘P0’ and ‘P1’. 

\subsection{{IL}}
When user follows procedure of applying transformation technique on the data, sending this data to LLM, getting transformed response, applying re transformation and obtaining final response, it is possible that user can lose some information in the response he obtains from the \gls{LLMs} due to provision of transformed data to \gls{LLMs} compared with the response user has otherwise obtained without any transformation. This can be understood with the example, suppose ‘Response1’ is the response user obtained from \gls{LLMs} without any transformation and ‘Response2’ is the re-transformed response user obtained from \gls{LLMs} using some transformation technique.  

Response1 = "According to our analysis, the company's revenue for the first quarter of 2023 increased \textbf{by 15\%} compared to the same period last year, reaching a  \textbf{total of \$10 million.} This growth was driven by a 20 \% increase in sales of our flagship product, which accounted for 60\% of the total revenue. However, operating expenses also increased \textbf{by 10\%}, mainly due to higher marketing and research and development costs. As a result, the company's net profit for the quarter was \textbf{\$1.2 million, a 12\% increase} from last year. Overall, the company's performance for the quarter was positive, but we recommend monitoring expenses closely to maintain profitability. " 

Response2= " According to our analysis, the company's revenue for the first quarter of 2023 increased compared to the same period last year. This growth was driven by a 20\% increase in sales of our flagship product, which accounted for 60\% of the total revenue. However, operating expenses also increased, mainly due to higher marketing and research and development costs. As a result, the company's net profit for the quarter was increased from last year. Overall, the company's performance for the quarter was positive, but we recommend monitoring expenses closely to maintain profitability. " 

As this can be seen from responses, Response2 lacks some of the important information like ‘company's revenue increased by 15\%’, ‘company’s revenue is of \$10 million’, ‘operating expenses also increased by 10\%’,’ company's net profit for the quarter was \$1.2 million, a 12\% increase’ which were there in Response1. So, this information which is lost from Response2 as compared with Response1 is called ‘Information loss.' This information can be crucial at many times so transformation techniques need to be chosen in such a way that the user will be having minimum information loss or information loss which can be tolerated. \gls{IL} is expressed in percentage like 20\% information loss. Two methods were used here for the calculation of the \gls{IL}. Information loss is expressed as  

\begin{center}
    \gls{IL}= (0.5*Manual information loss) + (0.5*Similarity based Information loss) 
\end{center}

\underline{\gls{ILS}:} This method uses hugging face sentence transformer model to calculate cosine similarity between our two responses. (e.g., Response 1 and Response 2). It converts the Responses into embedding and then compares how similar these sentences are based on cosine similarity and comes back with similarity score ranging from 0 to 1. \gls{ILS} is calculated as  

\begin{center}
    \gls{ILS}= 1 – Similarity score 
\end{center}

\pagebreak
\underline{\gls{ILM}:} One drawback of \gls{ILS} is it can give information loss more than 0\% even if both responses meanings are same. For example, if user provides context as “Mango is Fruit”, ask question as “What is Mango?” then one response can be “Mango is a fruit” and other response can be “A fruit”. Now both these responses are giving the right answer to the question and there is no information loss here, but \gls{ILS} can still give information loss greater than 0 as these two responses are not same and occupies different positions in the vector space when compared from cosine similarity. To overcome this, \gls{ILM} is also taken into consideration, adding human into loop. Manual Information loss is the loss in which human analyses both responses (e.g., Response1 and Response2), compares loss of information and then provides information loss ranging from 0 to 1. Here for comparison, a person can consider information which he/she finds important, it can be some figures, names or anything. Method of calculation of \gls{ILM} is given as 

\begin{center}
    \gls{ILM} =  Information loss from response obtained after use of Transformation Technique / Total important information in response obtained without use of Transformation Technique
\end{center}

Consider Response1 and Response2 to understand \gls{ILM}. So important information lost from Response2 are figures like “15\%”, “total of \$10 million”, “by 10\%”, “\$1.2 million” and “a 12\% increase”. So, this count is 5. Total important information in Response1 is “15\%”, “total of \$10 million”, “by 10\%”, “\$1.2 million” and “a 12\% increase”, “first quarter of 2023”, “20\% increase”, “flagship product”, “60\% of the total revenue” and count of this is 9. So, \gls{ILM} is 5/9 = 0.55

\subsection{Transformation cycle}
A complete process of applying Transformation Technique to data, sending this data to the \gls{LLMs}, obtaining transformed response from \gls{LLMs}, applying re-transformation on this response and obtaining final response is called as ‘Transformation cycle’. This term is used in paper for ease of understanding and to avoid describing this cycle again and again. 

\subsection{{STT}}
When Transformation cycle is applied using \gls{UPT} Transformation and suppose user has provided tokens for the words on which \gls{LLMs} are trained on, then it can be possible that \gls{LLMs} can use their trained understanding as a response, which means those are sensitive to some of tokens. This can be understood with one example, suppose user has given token ‘Mango’ for the word ‘Rose’.  Transformation cycle is as below. 

\begin{table}[h]
\centering
\begin{tabular}{|p{3in}|p{3in}|}
\hline
Original Text    & \textbf{Rose} is a flowering plant that is widely recognized for its beauty, fragrance, and symbolic significance. It belongs to the family Rosaceae and is native to Asia but is now cultivated in many parts of the world. \textbf{Rose} comes in a variety of colors, such as red, pink, yellow, and white, and is commonly used in gardens, bouquets, and various decorative arrangements.    \\ \hline
\gls{UPT} Transformed Text   & \textbf{Mango} is a flowering plant that is widely recognized for its beauty, fragrance, and symbolic significance. It belongs to the family Rosaceae and is native to Asia but is now cultivated in many parts of the world. \textbf{Mango} comes in a variety of colors, such as red, pink, yellow, and white, and is commonly used in gardens, bouquets, and various decorative arrangements.   \\ \hline

\end{tabular}
\end{table}

\begin{longtable}{|p{3in}|p{3in}|}
\hline
Question   & What is \textbf{Rose?}  \\ \hline
\gls{UPT} Transformed question along with transformed context   & The \textbf{Mango} is a flowering plant that is widely recognized for its beauty, fragrance, and symbolic significance. It belongs to the family Rosaceae and is native to Asia but is now cultivated in many parts of the world. \textbf{Mango} comes in a variety of colors, such as red, pink, yellow, and white, and is commonly used in gardens, bouquets, and various decorative arrangements. What is \textbf{Mango?}   \\ \hline
Transformed Response from \gls{LLMs}   & I'm sorry, but the information you have provided about \textbf{Mango} is not accurate. \textbf{Mango} is not a member of the Rosaceae family; it belongs to the Anacardiaceous family. \textbf{Mango} is also not typically used in gardens or decorative arrangements, as it is a fruit that is typically eaten fresh or used in cooking. \textbf{Mango} comes in a variety of colors, including green, yellow, and orange, but not pink or white. \textbf{Mango} trees do produce flowers, but the fruit is the most well-known and widely used part of the plant.  \\ \hline
Final Re-transformed Response  & I'm sorry, but the information you have provided about \textbf{Rose} is not accurate. \textbf{Rose} is not a member of the Rosaceae family; it belongs to the Anacardiaceous family. \textbf{Rose} is also not typically used in gardens or decorative arrangements, as it is a fruit that is typically eaten fresh or used in cooking. \textbf{Rose} comes in a variety of colors, including green, yellow, and orange, but not pink or white. \textbf{Rose} trees do produce flowers, but the fruit is the most well-known and widely used part of the plant.   \\ \hline
\end{longtable}

So, user has provided information that ‘Mango is Flower’. This is user provided knowledge, this information can conflict with the information of ‘Mango is fruit’ on which \gls{LLMs} are already trained on. When user asks to \gls{LLMs} that ‘what is Mango?’  then it gives response as ‘Mango is fruit’ so it means that even if user has provided information that ‘Mango is flower’, still it is using data on which it is pre-trained on which means that it is sensitive to this kind of information. So likewise, \gls{LLMs} can be sensitive to many types of words and will give a different response than expected, so this is called \gls{STT}, and this should be considered into effect before giving tokens otherwise user will be getting responses which will come from \gls{LLMs} trained understanding rather than information user is providing to it. Also, \gls{LLMs} can give out of context answers from data on which it is already trained on. \gls{STT} is expressed as either of these formats like yes or no, 100\% or 0\%, 1 or 0. \gls{STT} of 100\% or 1 means \gls{LLMs} are sensitive to Transformation Technique and will use its pre-trained knowledge than using our provided information to generate the response.

\subsection{Prompt Engineering}

Prompt engineering refers to the process of designing effective prompts or input examples that help \gls{LLMs} learn to perform a specific task. It can also be seen as a way of communicating more effectively with \gls{LLMs}, such that the resulting output adheres more closely to the context and constraints within which the problem needs to be addressed.

This is a need more than a good to have, as \gls{LLMs} are prone to hallucinating - generating real looking responses that are factually inaccurate. As part of our technique, we use Prompt Engineering for two main reasons:

\begin{itemize}
  \item To act as a constraint, making the output relevant to the context - \gls{LLMs} give out of context answers in their vanilla state.
  \item To return a pre-defined response if the question goes out of context - We do this by POS Transformation-prompting, which means adding an instructional prompt at the end of each context and question pair. There are better ways to do this, such as prompt evaluation, which would be beyond the scope of discussion in this paper. 
\end{itemize}
\raggedbottom
\pagebreak

\section{Experimental Results}
We have compared responses from \gls{LLMs} for different combinations of \gls{UPT} Transformation, \gls{NER} Transformation and \gls{PoS} Transformation. In different combinations, Transformation Technique is used alone, or different stages are applied like performing other Transformation Technique on top of the first Transformation Technique. As an example, \gls{UPT}+\gls{NER} means first \gls{UPT} Transformation is applied and then \gls{NER} Transformation. At the time of re-transformation also the same stages of re-transformation will be applied to get the clarified response. Like first \gls{NER} re-transformation will be done and then \gls{UPT} re-transformation to obtain response. 
We obtained responses for three kinds of questions.  

\begin{enumerate}
  \item Pointed questions: These are straightforward questions having one- or two-line answers. Sample question will be ‘who is owner of Facebook?’ and answer will be ‘The owner of Facebook is Mark Zuckerberg.’  
  \item Key questions: These are more complex questions having more than 2 lines of answers. Example question will be ‘What are key accomplishments.
  \item Summarizing questions: These are questions for summarizing reports. Example question will be ‘Summarize this report’.  
\end{enumerate}

We asked 40 questions belonging to the above three kinds to LLM (LLM considered here is Chat-GPT), calculated \gls{STT}, \gls{ILM}, \gls{ILS} and \gls{IL} for all these responses and calculated average of these measures for each Transformation Technique and these are presented in Table 1.1. 

\begin{table}[h]
\centering
\resizebox{\textwidth}{!}{%
\begin{tabular}{|p{1.6in}|p{1.3in}|p{1.3in}|p{1.3in}|p{1.3in}|}
\hline
\textbf{Transformation Technique}  & \textbf{\gls{STT}}  &\textbf{\gls{ILM}}   & \textbf{\gls{ILS}}  &\textbf{\gls{IL}}   \\ \hline
 \gls{UPT} &0.00\%   &1.28\%   &12.70\%   &6.99\%   \\ \hline
 \gls{NER} &7.69\% &35.90\% &16.40\% &26.15\%   \\ \hline
 \gls{PoS} & 2.56\% &13.85\% &14.12\% &13.98\%   \\ \hline
 \gls{UPT}+\gls{NER} &2.56\% &28.97\% &27.94\% &28.46\% \\ \hline
 \gls{UPT} + \gls{PoS} &0.00\% &22.82\% &19.95\% &21.39\% \\ \hline
 \gls{NER} + \gls{PoS} &5.13\% &32.95\% &26.72\% &29.84\%   \\ \hline
 \gls{UPT} + \gls{NER} + \gls{PoS} &0.00\% &43.08\% &33.80\% &38.44\%   \\ \hline
\end{tabular}%
}
\end{table}

\begin{center}
    Table 1.1: Experiment Result for questions 
\end{center}

From Table 1.1, \gls{IL} is observed minimum for the \gls{UPT} and maximum for the \gls{UPT}+\gls{NER}+\gls{PoS}. This seems obvious as it has more stages of Transformation Technique applied over it. All Other techniques give \gls{IL} in similar range. From the above information losses which were obtained based on the final responses from \gls{LLMs}, it is observed that we are outperforming traditional data perturbation techniques in terms of both utility preservation and privacy protection as \gls{LLMs} has given these responses based on our Transformation Technique. Also, \gls{STT} is observed in Transformation Technique like \gls{NER}, \gls{PoS}, \gls{UPT}+\gls{NER} and \gls{NER}+\gls{PoS}. It means that for these techniques for few of questions, \gls{LLMs} was sensitive or has given answer from its data on which it was pre-trained on.  

Table 1.2 shows \gls{IL} and \gls{STT} for the final responses for questions when Prompt Engineering is used. 

\begin{table}[h]
\centering
\resizebox{\textwidth}{!}{%
\begin{tabular}{|p{1.6in}|p{1.3in}|p{1.3in}|p{1.3in}|p{1.3in}|}
\hline
\textbf{Transformation Technique}  & \textbf{\gls{STT}}  &\textbf{\gls{ILM}}   & \textbf{\gls{ILS}}  &\textbf{\gls{IL}}   \\ \hline
 \gls{UPT} &0.00\%   &1.28\%   &12.70\%   &6.99\%   \\ \hline
 \gls{NER} &0.00\% &34.21\% &23.62\% &28.48\%   \\ \hline
 \gls{PoS} &2.56\% &13.85\% &14.12\% &13.98\%   \\ \hline
 \gls{UPT} + \gls{NER} &0.00\% &28.97\% &29.82\% &29.39\% \\ \hline
 \gls{UPT} + \gls{PoS} &0.00\% &22.82\% &19.95\% &21.39\% \\ \hline
 \gls{NER} + \gls{PoS} &0.00\% &32.95\% &31.14\% &32.04\%   \\ \hline
 \gls{UPT} + \gls{NER} + \gls{PoS} &0.00\% &43.08\% &33.80\% &38.44\%   \\ \hline
\end{tabular}%
}
\end{table}

\begin{center}
    Table 1.2: Experiment Result for questions with Prompt Engineering 
\end{center}

As it can be observed that Prompt Engineering has reduced \gls{STT} to zero almost for all the techniques.  
\raggedbottom
\pagebreak

\section{Conclusion}
We have provided and tested Life of \gls{PII} for sending information to \gls{LLMs} ensuring protection of critical information and keeping the semantic meaning thereby ensuring appropriate responses from the \gls{LLMs}. Selection of Transformation Technique to use depends on the type of use case. For use cases where only a few terms/information needs to be protected and requires minimum information loss, \gls{UPT} can be used. Use cases where protecting most of the information is a top priority and information loss is on less priority, \gls{UPT}+\gls{NER}+\gls{PoS} can be used. If balance of both protecting information as well as less information loss is required, then other Transformation Technique can be used. Also, to ensure \gls{LLMs} give response within context, prompt engineering can be used. Currently in \gls{UPT}, tokens are manually provided by the user to have full control over the technique which can make the process a little slower. In the future we will work on automating token provision process so that it can make process faster and remove task of user to add appropriate tokens.

\bibliographystyle{unsrt}
\bibliography{references}

\end{document}